\documentclass[runningheads]{llncs}
\usepackage{graphicx}
\usepackage{comment}
\usepackage{amsmath,amssymb} % define this before the line numbering.
\usepackage{color}

\usepackage[width=122mm,left=12mm,paperwidth=146mm,height=193mm,top=12mm,paperheight=217mm]{geometry}

\usepackage{epsfig}
\usepackage{graphicx}        
\usepackage{amsmath}
\usepackage{amssymb}
\usepackage{color}
\usepackage{tabularx}
\usepackage{booktabs}
\usepackage{multirow}
\usepackage[linesnumbered,ruled]{algorithm2e}
\usepackage{array}
\usepackage{xspace}
\usepackage{balance}
\usepackage{enumitem}
\usepackage{mathtools}
\usepackage{rotating}
\usepackage{url}
\usepackage{tabulary}
\usepackage[table]{xcolor}
\usepackage{subcaption}
\usepackage[export]{adjustbox}
\usepackage{ctable}
\usepackage{diagbox}
\usepackage{subcaption}
\usepackage{makecell}
\usepackage{amssymb}
\usepackage[linesnumbered,ruled]{algorithm2e}

\newcommand{\ie}{\textit{i}.\textit{e}.}
\newcommand{\eg}{\textit{e}.\textit{g}.}
\newcommand{\etal}{\textit{et al}.}

\DeclareMathOperator*{\argmax}{arg\,max}

\graphicspath{{./figs/}}

\usepackage[pagebackref=true,breaklinks=true,letterpaper=true,colorlinks,bookmarks=false]{hyperref}

\begin{document}
	% \linenumbers
	\pagestyle{headings}
	\mainmatter
	\def\ECCVSubNumber{}  % Insert your submission number here
	
	\title{Towards Noise-resistant Object Detection with Noisy Annotations}
	
	%\begin{comment}
	\titlerunning{} 
	\authorrunning{} 
	\author{Junnan Li \and
	Caiming Xiong \and
	Richard Socher \and
	Steven Hoi}
    \institute{Salesforce Research}
	%\end{comment}
	%******************
	
	\begin{comment}
	\titlerunning{Abbreviated paper title}
	\author{First Author\inst{1}\orcidID{0000-1111-2222-3333} \and
	Second Author\inst{2,3}\orcidID{1111-2222-3333-4444} \and
	Third Author\inst{3}\orcidID{2222--3333-4444-5555}}
	%
	\authorrunning{F. Author et al.}
	\institute{Princeton University, Princeton NJ 08544, USA \and
	Springer Heidelberg, Tiergartenstr. 17, 69121 Heidelberg, Germany
	\email{lncs@springer.com}\\
	\url{http://www.springer.com/gp/computer-science/lncs} \and
	ABC Institute, Rupert-Karls-University Heidelberg, Heidelberg, Germany\\
	\email{\{abc,lncs\}@uni-heidelberg.de}}
	\end{comment}
	%******************
	\maketitle
	
	\begin{abstract}

Training deep object detectors requires significant amount of human-annotated images with accurate object labels and bounding box coordinates,
which are extremely expensive to acquire. 
Noisy annotations are much more easily accessible,
but they could be detrimental for learning.
We address the challenging problem of training object detectors with noisy annotations,
where the noise contains a mixture of label noise and bounding box noise. 
We propose a learning framework which jointly optimizes object labels, bounding box coordinates, and model parameters by performing alternating noise correction and model training.
To disentangle label noise and bounding box noise,
we propose a two-step noise correction method.
The first step performs class-agnostic bounding box correction by minimizing classifier discrepancy and maximizing region objectness.
The second step distils knowledge from dual detection heads for soft label correction and class-specific bounding box refinement.
We conduct experiments on PASCAL VOC and MS-COCO dataset with both synthetic noise and machine-generated noise.
Our method achieves state-of-the-art performance by effectively cleaning both label noise and bounding box noise.
Code to reproduce all results will be released.
	
\end{abstract}
	\section{Introduction}
\label{sec:introduction}

The remarkable success of modern object detectors % such as Faster R-CNN~\cite{faster_rcnn} and Mask R-CNN~\cite{mask_rcnn} 
largely relies on the collection of large-scale datasets such as ILSVRC~\cite{imagenet} and MS-COCO~\cite{mscoco} with extensive bounding box annotations.
However, it is extremely expensive and time-consuming to acquire high-quality human annotations with accurate object labels and precise bounding box coordinates.
For example, annotating each bounding box in ILSVRC requires 42s on Mechanical Turk using a technique specifically developed for efficient annotation~\cite{Crowdsourcing}.
%For example,
%8 humans were required to annotate each image in MS-COCO,
On the other hand,
there are ways to acquire bounding box annotations at lower costs,
such as limiting the annotation time, reducing the number of crowd-workers, or utilizing machine-generated annotations.
However,
these methods would inevitably yield annotations with both \textit{label noise} (\ie~wrong object classes) and \textit{bounding box noise} (\ie~inaccurate object locations),
which could be detrimental for learning. 

Learning with label noise has been an active area of research.
Existing methods primarily take a label/loss correction approach.
Some methods perform label correction using the predictions from the model and modify the loss accordingly~\cite{Reed_2015_ICLR,Tanaka_CVPR_2018}.
Other methods treat samples with small loss as those with clean labels,
and only allow clean samples to contribute to the loss~\cite{mentornet,co-teaching}.
However,
most of those methods focus on the image classification task where the existence of an object is guaranteed.

\begin{figure}[!t]
 \centering
   \includegraphics[width=0.98\linewidth]{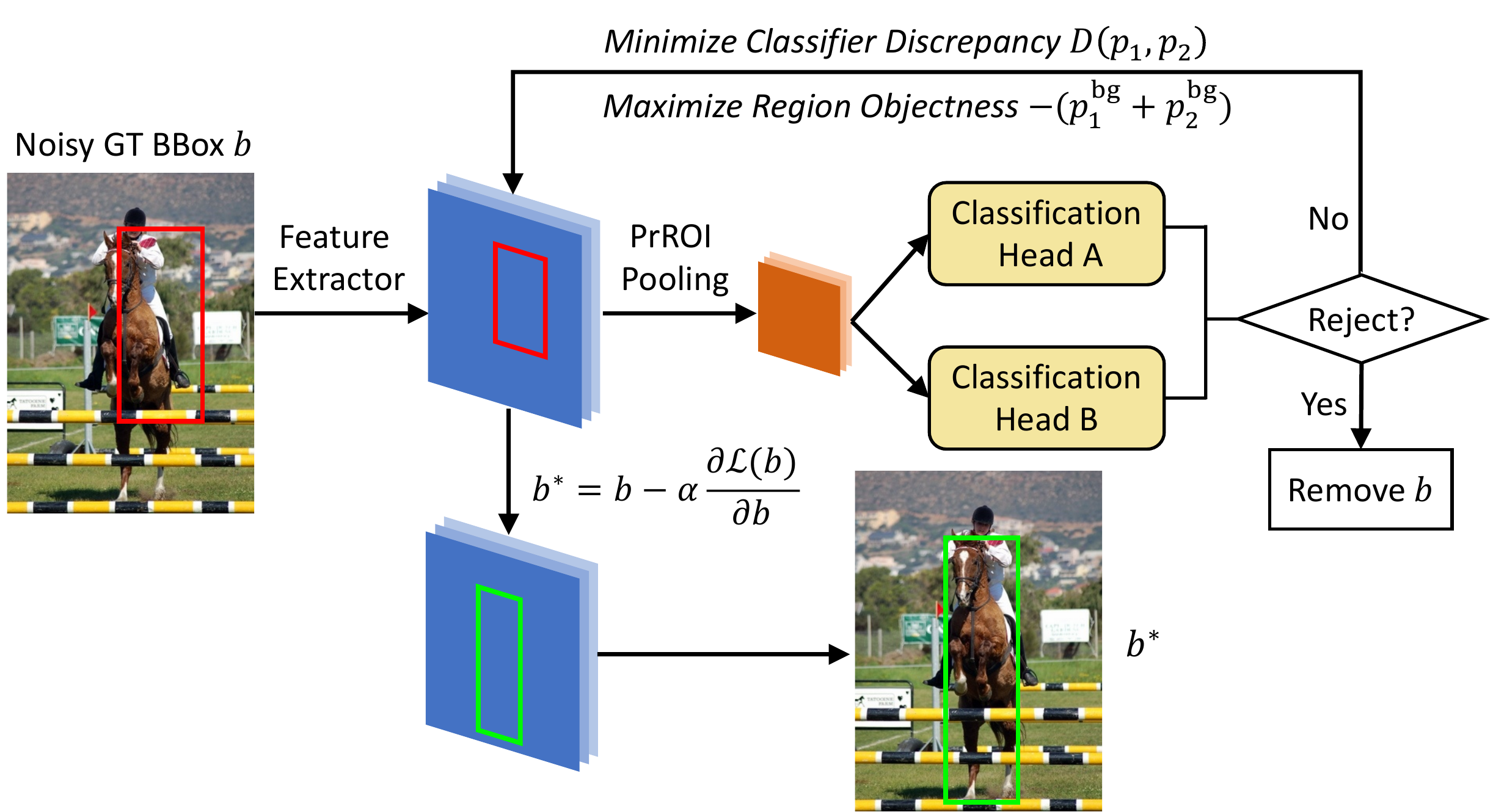}
   \caption
  	{
  	\small
		Our Class-Agnostic Bounding Box Correction (CA-BBC) disentangles bounding box (bbox) noise from label noise, by directly optimizing the noisy bbox coordinates regardless of its class label. We use two diverged classifiers to predict the same image region, and update the bbox $b$ to $b^*$ by minimizing classifier discrepancy and maximizing region objectness. Boxes with very low objectness are rejected as false positives. 		
	 } 
	 \vspace{-3ex}
  \label{fig:CA}
 \end{figure}

Several recent works have studied object detection with noisy annotations.
Zhang~\etal~\cite{noisy_AAAI} focus on the weakly-supervised (WS) setting where only image-level labels are available,
and find reliable bounding box instances as those with low classification loss.
Gao~\etal~\cite{noisy_ICCV} study a semi-supervised (SS) setting where the training data contains a small amount of fully-labeled bounding boxes and a large amount of image-level labels,
and propose to distil knowledge from a detector pretrained on clean annotations.
These WS or SS object detection methods all require some clean annotations to be available.

In this work,
we address a more challenging and practical problem,
where the annotation contains \textit{an unknown mixture} of label noise and bounding box noise.
Furthermore, \textit{we do not assume access to any clean annotations}.
The entanglement of label noise and bounding box noise increases the difficulty to perform noise correction.
A commonly used noise indicator, namely the classification loss,
is incapable to distinguish label noise from bounding box noise,
because both wrong class labels and inaccurate box locations could cause samples to have larger loss.
Furthermore, 
it is problematic to correct noise directly using the model predictions,
because label correction requires accurate bounding box coordinates to crop the object,
whereas class-specific bounding box regression requires true class labels to choose the correct regression offset.

To overcome these difficulties,
we propose a two-step noise correction procedure.
In the first step,
we perform class-agnostic bounding box correction (CA-BBC),
which seeks to decouple bounding box noise from label noise,
and optimize the noisy ground-truth (GT) bounding box regardless of its class label.
%by using two diverged classification heads.
An illustration of CA-BBC is shown in Figure~\ref{fig:CA}.
It is based on the following intuition:
if a bounding box tightly covers an object,
then two diverged classifiers would agree with each other and produce the same prediction.
Furthermore,
both classifiers would have low scores for the background class, \ie, high objectness scores.
Therefore,
we directly regress the noisy GT bounding box to minimize both classifier discrepancy and background scores.
CA-BBC also has the option to reject a bounding box as false positive if the objectness score is too low.

In the second step,
we leverage the model's output for label noise correction and class-specific bounding box refinement.
It has been shown that co-training two models can filter different types of noise and help each other learn~\cite{co-training,co-teaching,Yu_ICML_2019,noisy_arxiv}.
Therefore,
we distil knowledge from the ensemble of dual detection heads for noise correction,
by generating soft labels and bounding box offsets.
We show that soft labels with well-adjusted temperature lead to better performance,
even for a clean dataset with no annotation noise.

To summarize,
this paper proposes a noise-resistant learning framework to train object detectors with noisy annotations,
where label noise and bounding box noise are entangled.
The proposed framework jointly optimizes object labels, bounding box coordinates, and model parameters by performing alternating noise correction and model training.
We conduct experiments on two popular benchmarks: PASCAL VOC and MS-COCO.
The model is trained on different levels of synthetic noise as well as machine-generated noise.
The proposed method outperforms state-of-the-art methods by a large margin,
achieving significant performance improvement under high noise level.
We also provide qualitative results to demonstrate the effectiveness of the proposed two-step noise correction,
and ablation studies to examine the effect of each component.

	\section{Related Work}
\label{sec:literature}

%This section reviews previous literatures in learning with noisy annotations for image classification and object detection tasks.
%Compared to previous works,
%this paper addresses a different and more challenging problem where label noise and bounding box noise are entangled.
%Note that the problem we address is also different from adversarial attacks~\cite{Zhang_2019_ICCV,Xie_2019_ICCV}.

\subsection{Crowdsourcing for Object Detection}
Crowdsourcing platforms such as Amazon Mechanical Turk (AMT) have enabled the collection of large-scale datasets.
Due to the formidable cost of human annotation,
many efforts have been devoted to effectively leverage the crowd and reduce the annotation cost.
However, even an efficient protocol designed to produce high-quality bounding boxes still report 42.4s to annotate one object in an image~\cite{Crowdsourcing}:
25.5s for drawing box and 16.9s for verification.
Other methods have been proposed which trade off annotation quality for lower cost,
by using click supervision~\cite{click},
human-in-the-loop labeling~\cite{human_machine,human_verification,dialog},
or exploiting eye-tracking data~\cite{eyetrack}.
These methods focus on reducing human effort,
rather than combating the annotation noise as our method does.

\subsection{Learning with Label Noise}

It has been shown that Deep Neural Networks (DNNs) can easily overfit to noisy labels in the training data,
leading to poor generalization performance~\cite{Zhang_ICLR_2017}.
Many works have addressed learning with label noise.
Some approaches correct noise by relabeling the noisy samples,
where noise is modeled through directed graphical models~\cite{Tong_CVPR_2015},
Conditional Random Fields~\cite{Vahdat_NIPS_2017},
or DNNs~\cite{Andreas_CVPR_2017,Lee_CVPR_2018}.
However,
these approaches rely on a small set of clean samples for noise correction.
Recently,
iterative relabeling methods have been proposed~\cite{Tanaka_CVPR_2018,Yi_2019_CVPR} which predict hard or soft labels using the model predictions.
Some approaches seek to correct noise by directly modifying the loss.
The bootstrapping method~\cite{Reed_2015_ICLR} introduces the model prediction into the loss,
which is later improved by~\cite{Ma_ICML_2018} through exploiting the dimensionality of feature subspaces. 
Other approaches filter noise by reweighting or selecting training samples~\cite{mentornet,Ren_ICML_2018,Chen_ICML_2019,Arazo_ICML_2019}.
Since DNNs learn clean samples faster than noisy ones, 
samples with smaller classification loss are usually considered to be clean~\cite{Arpit_ICML_2017}.
To avoid error accumulation during the noise correction process,
co-teaching~\cite{co-teaching} trains two networks simultaneously,
where each network selects small-loss samples to train the other.
Co-teaching$+$~\cite{Yu_ICML_2019} further keeps the two networks diverged by training on disagreement data.

%All of the above methods focus on the classification problem,
%where an object's existence in the image is guaranteed. 

\subsection{Weakly-supervised Object Detection}

Weakly-supervised object detection aims to learn object detectors with only image-level labels.
Most existing works formulate it as a multiple instance learning (MIL) task~\cite{MIL},
where each label is assigned to a bag of object proposals.
A common pipeline is to iteratively alternate between mining object instances using a detector and training the detector using the mined instances~\cite{Deselaers_ECCV_2010,Cinbis_PAMI_2017}.
%A two-stream end-to-end network is designed which outputs proposal scores by multiplying scores from the classification stream and the localization stream.
To address the localization noise in the object proposals,
Zhang~\etal~\cite{noisy_AAAI} propose an adaptive sampling method which finds reliable instances as those with high classification scores,
and use the reliable instances to impose a similarity loss on noisy images.
Different from weakly-supervised object detection which assumes that the correct object label is given,
our method deals with label noise and bounding box noise at the same time.

\begin{figure*}[!t]
 \centering
   \includegraphics[width=\linewidth]{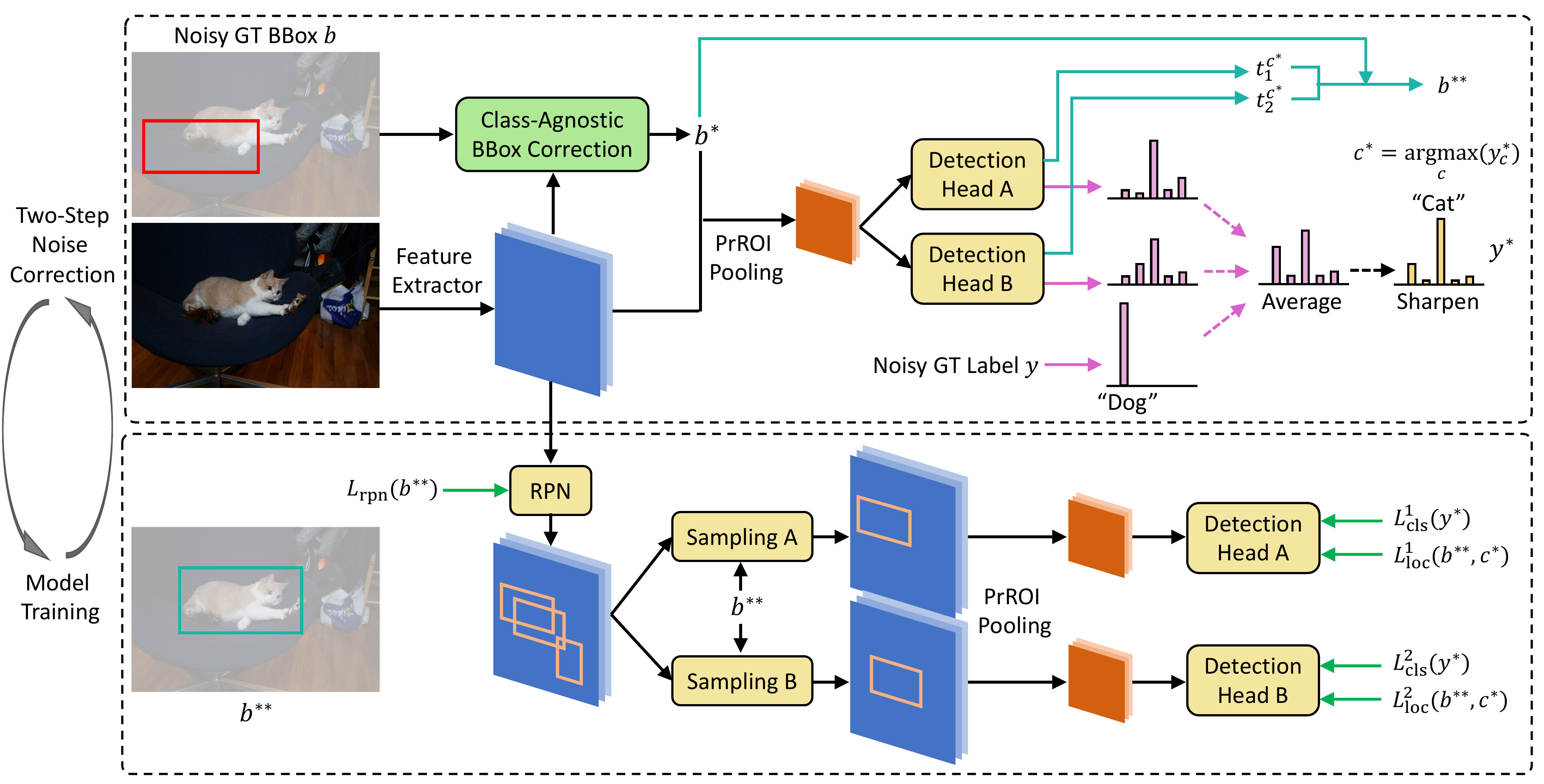}
   \vspace{-3ex}
   \caption
  	{
  	\small
		The proposed framework towards noise-resistant object detection from noisy annotations, which alternately performs noise correction (with fixed model parameters) and model training (with corrected annotations) for each mini-batch.
	    The noise correction procedure consists of two steps:
	    (1) the class-agnostic bounding box correction (Figure~\ref{fig:CA}) disentangles bbox noise and label noise; (2) the class-specific correction step uses dual detection heads to generate soft labels for label correction and refine bounding boxes using class-specific bbox offsets. 
	    %The updated annotations are then used to train the model.
	    The two detection heads are kept diverged by different initialization and different random RoI sampling during training.
} 
 \vspace{-1ex}
  \label{fig:framework}
 
 \end{figure*}
\subsection{Semi-supervised Object Detection}
Semi-supervised methods train object detectors using training data with bounding box annotations for some images and only image-level labels for other images~\cite{LSDA_semi,Tang_semi,Jasper_semi,noisy_ICCV}.
A recent method also incorporates unlabeled images~\cite{Jisoo_semi}.
Gao~\etal~\cite{noisy_ICCV} proposes an iterative training-mining framework consisting of detector initialization, box mining, and detector retraining. 
To address the annotation noise of the mined boxes,
they use a detector pretrained on clean annotations for knowledge distillation.
Different from all semi-supervised learning methods, our method does not need access to any clean annotations.

	\section{Method}
\label{sec:method}

\subsection{Overview}

Given a training dataset with images $\mathcal{X}$,
noisy object labels $\mathcal{Y}$,
and noisy bounding boxes $\mathcal{B}$,
our method aims to train an object detector parameterized by $\Theta$, by jointly optimizing $\mathcal{Y}$, $\mathcal{B}$ and $\Theta$. We first warm-up $\Theta$ where we train the detector in a standard manner using the original noisy annotations.
After the warm-up,
we perform alternating optimization on the annotations and the model.
Specifically,
for each mini-batch of data $X=\{x_i\}$, $Y=\{y_i\}$, $B=\{b_i\}$,
we first keep $\Theta$ fixed and perform noise correction to update $Y$ and $B$,
then we used the corrected annotations to update $\Theta$.
An overview of the algorithm is shown in Algorithm~\ref{alg:framework}.

We use a popular two-stage object detector (\eg~Faster-RCNN~\cite{faster_rcnn}),
which consists of a backbone feature extractor parameterized by $\theta_\mathrm{cnn}$, a Region Proposal Network (RPN) $\theta_\mathrm{rpn}$, a classification head $\theta_c$, and a bounding box (bbox) regression head $\theta_b$.
Note that $\theta_c$ and $\theta_b$ have shared layers.
Let \textit{detection head} with parameters $\theta_d$ denote the union of the classification head and the bbox regression head.
During training,
we simultaneously train two detection heads $\theta_d^1=\{\theta_c^1,\theta_b^1\}$ and $\theta_d^2=\{\theta_c^2,\theta_b^2\}$,
%with shared backbone.
which are kept diverged from each other by different parameter initializations and different training instance (\ie~RoI) sampling.
The dual detection heads are utilized to correct annotation noise.

\newlength{\textfloatsepsave} 
\setlength{\textfloatsepsave}{\textfloatsep}
\setlength{\textfloatsep}{0pt}
\begin{algorithm}[!t]

	\SetKw{KwIn}{in}
	\DontPrintSemicolon
	
	\textbf{Input:} model $\Theta=\{\theta_\mathrm{cnn},\theta_\mathrm{rpn},\theta_d^1,\theta_d^2\}$, noisy training dataset $(\mathcal{X},\mathcal{Y},\mathcal{B})$.\\

	\While{$\mathrm{not~MaxIters}$}
	{	
	
	Mini-batch $X=\{x_i\}$, $Y=\{y_i\}$, $B=\{b_i\}$. \\

	\For {$b$ \KwIn $B$}    
    {
		Update $b\rightarrow b^*$  with CA-BBC (Eq.~\ref{eqn:Lb} \&~\ref{eqn:b*}).
	}
 
    \For {$(y,b^*)$ \KwIn $(Y,B^*)$}   
    {
    	Update $y\rightarrow y^*$  with dual-head soft label correction (Eq.~\ref{eqn:bootstrap} \&~\ref{eqn:sharp}).\\
    	Update $b^*\rightarrow b^{**}$  with class-specific bbox refinement (Eq.~\ref{eqn:bbox}).
    }
    
    Update $\Theta$ by SGD on  $L_\mathrm{rpn}(B^{**})$,  $L_\mathrm{cls}^{1+2}(Y^*)$, $L_\mathrm{loc}^{1+2}(B^{**},Y^*)$.\\
    %Update dual-teacher by EMA (Eq.~\ref{eqn:ema}).
	}
	\caption{alternating two-step noise correction and model training.}
	\label{alg:framework}
\end{algorithm}
\setlength{\textfloatsep}{\textfloatsepsave}

Due to the entanglement of an unknown mixture of label noise and bbox noise,
it is difficult to correct both types of annotation noise in a single step.
Therefore,
we propose a two-step noise correction method.
In the first step,
we perform class-agnostic bounding box correction (CA-BBC),
which disentangles bbox noise and label noise by directly optimizing the noisy GT boxes regardless of their class labels.
In the second step, we utilize the outputs from dual detection heads for label noise correction and class-specific bbox refinement.
Figure~\ref{fig:framework} shows an illustration of our framework. 
Next we delineate the details.

\subsection{Class-agnostic Bounding Box Correction}

We first correct bounding box noise by updating $B\rightarrow B^*$ regardless of the label noise in $Y$.
As illustrated in Figure~\ref{fig:CA},
CA-BBC uses two diverged classification heads to produce two sets of class predictions on the same image region,
and updates the bounding box to \textit{minimize classifier discrepancy} and \textit{maximize region objectness}.
The intuition is:
if a bounding box tightly covers an object,
then two classifiers would agree with each other and produce the same predictions.
Moreover,
both predictions would have low scores on the background class.

Specifically, given an image $x \in X$,
the backbone first extracts a convolutional feature map.
For each noisy GT bounding box $b \in B$,
we perform a RoI-Pooling operation on the feature map to extract a fixed-sized feature $\phi(x,b)$.
Then we give the RoI feature to the two classification heads to produce two sets of softmax predictions over $C+1$ classes (including the background class), $p_1(\phi(x,b);\theta_c^1)$ and $p_2(\phi(x,b);\theta_c^2)$.
For simplicity we denote them as $p_1$ and $p_2$.
The discrepancy between the two predictions is defined as their L2 distance:
\begin{equation}
\label{eqn:dis}
	\mathcal{D}(p_1,p_2)=\left\lVert p_1-p_2 \right\rVert^2_2.
\end{equation}

Minimizing the classifier discrepancy \textit{w.r.t} the bounding box will push it to a region where the two classifiers agree on its class label.
To prevent the bounding box from simply moving to a background region,
we also minimize the classifiers' scores on the background class, $p_1^\mathrm{bg}$ and $p_2^\mathrm{bg}$.
In other words,
we want to maximize the objectness of the region covered by the bounding box.

Therefore, we aim to find the optimal $b^*$ that minimizes the following objective function:
\begin{equation}
\label{eqn:Lb}
\mathcal{L}(b)=\mathcal{D}(p_1,p_2)+\lambda(p_1^\mathrm{bg}+p_2^\mathrm{bg}),
\end{equation}
where $\lambda$ controls the balance of the two terms and is set to 0.1 in our experiments.

For faster speed,
we estimate $b^*$ by performing a single step of gradient descent to update $b$:
\begin{equation}
\label{eqn:b*}
b^* = b-\alpha \frac{\partial\mathcal{L}(b)}{\partial b} ,
\end{equation}
where $\alpha$ is the step size.

Since RoI-Pooling~\cite{faster_rcnn} or RoI-Align~\cite{mask_rcnn} performs discrete sampling on the feature map to generate $\phi(x,b)$,
$\mathcal{L}(b)$ is not differentiable w.r.t $b$.
Therefore,
we adopt the Precise RoI-Pooling method~\cite{prroi},
which avoids any quantization of coordinates and has a continuous gradient on $b$.

We also observe that the entropy of the classifiers' predictions over object classes would decrease after updating $b$ to $b^*$.
A lower entropy suggests that the classifiers are more confident of their predicted object class~\cite{confidence},
which verifies our assumption that $b^*$ contains more representative information for one and only one object.

In order to handle false positive bboxes that do not cover any object,
we add a reject option which removes $b$ from the ground-truth if both classifiers give a low objectness score (high background score), 
$p_1^\mathrm{bg}>0.9$ and $p_2^\mathrm{bg}>0.9$.
 
\subsection{Dual-Head Distillation for Noise Correction}
\label{sec:dual-head}
In the second step of noise correction,
we perform class-specific self-distillation for label noise correction and bbox refinement.
Inspired by the co-training methods~\cite{co-training,co-teaching,Yu_ICML_2019},
we simultaneously train two diverged detection heads which can filter different types of noise,
and distil knowledge from their ensemble to clean the annotation noise.
Using the ensemble of two heads helps alleviate the confirmation bias problem (\ie~a model confirms its own mistakes~\cite{Tarvainen_NIPS_17}) that commonly occurs in self-training and achieves better robustness.
%we utilize the outputs from the dual detection heads for label noise correction and class-specific bbox refinement.
%It has been shown in~\cite{MLNT} that the mean-teacher model~\cite{Tarvainen_NIPS_17} is more resistant to label noise at early stage of training and can be used to guide the student model.
%It has been shown that simultaneously training two models can filter different types of label noise~\cite{co-training,co-teaching,Yu_ICML_2019}.
%We propose a dual-head noise correction method,
%which leverages the outputs the two detection heads.
%we simultaneously train two diverged heads with distinct abilities to filter different types of noise,
%and use their ensemble to clean the annotation noise.
%Our method distills knowledge from each detection head to teach the other,
%which helps alleviate the confirmation bias problem (\ie~a model confirms its own mistakes~\cite{Tarvainen_NIPS_17}) and achieves robustness to noise.

\noindent\textbf{Soft label correction.}
Given the RoI feature $\phi(x,b^*)$,
the two classification heads produce two sets of softmax predictions over object classes,
$p_1^*$ and $p_2^*$.
%$p^\mathrm{tch}_1$ and $p^\mathrm{tch}_2$.
Inspired by the bootstrapping method~\cite{Reed_2015_ICLR},
we use the classifiers' predictions to update the noisy GT label.
Let $y\in\{0,1\}^C$ represent the GT label as a one-hot vector over $C$ classes,
we create the soft label by first averaging the classifiers' predictions and the GT label:
\begin{equation}
\label{eqn:bootstrap}
\bar{y}=(p_1^*+p_2^*+y) \big/3.
\end{equation}
Then we apply a sharpening function on the soft label to reduce the entropy of the label distribution.
The sharpening operation is defined as:
\begin{equation}
\label{eqn:sharp}
\small
y^*={\bar{y}^{c\frac{1}{T}}} \bigg/ \sum_{c=1}^C {\bar{y}^{c\frac{1}{T}}},~ c=1,2,...,C,
\end{equation}
where $\bar{y}^c$ is the score for class $c$.
The temperature $T$ controls the `softness' of the label and is set to 0.4 in our experiments.
A lower temperature decreases the softness and has the implicit effect of entropy minimization,
which encourages the model to produce high confidence predictions and allows better decision boundary to be learned~\cite{Grandvalet_NIPS_2005,mixmatch}.

\noindent\textbf{Class-specific bounding box refinement.}
The two bbox regression heads produce two sets of per-class bounding box regression offsets,
$t_1$ and $t_2$.
Let $c^*$ denote the class with the highest score in the soft label,
\ie~$c^*=\argmax_c y^*_c,c=1,2,...,C.$
We refine the bounding box $b^*$ by merging the class-specific outputs from both bbox regression heads:
\begin{equation}
\label{eqn:bbox}
\begin{split}
	t &= (t_1^{c^*}+t_2^{c^*})/2 \\
	b^{**} &=b^*+\rho t,
\end{split}
\end{equation}
where $t_1^{c^*}$ and $t_2^{c^*}$ are the bounding box offsets for class $c^*$,
and $\rho$ controls the magnitude of the refinement.

\vspace{-2ex}
\subsection{Model Training}
Let $Y^*$ and $B^{**}$ denote a mini-batch of soft labels and refined bounding boxes,
respectively.
We use them as the new GT to train the model.
Specifically,
we update $\Theta=\{\theta_\mathrm{cnn},\theta_\mathrm{rpn},\theta_d^1,\theta_d^2\}$ to optimize the following losses:
(1) the loss function of RPN defined in~\cite{faster_rcnn}, {\small$L_\mathrm{rpn}(B^{**})$};
(2) the classification loss for the two detection heads, {\small$L_\mathrm{cls}^1(Y^*)$ and $L_\mathrm{cls}^2(Y^*)$}, defined as the cross-entropy loss {\small$\sum_i -y_i^*\log(p_i)$};
(3) the localization loss for the two detection heads, {\small$L_\mathrm{loc}^1(B^{**},Y^*)$} and {\small$L_\mathrm{loc}^2(B^{**},Y^*)$,}
defined as the smooth L1 loss~\cite{fast-rcnn}.

%\begin{itemize}
%\vspace{-0.5ex}
%\item[(1)] the loss function of RPN defined in~\cite{faster_rcnn}, {\small$L_\mathrm{rpn}(B^{**})$.}
%\item[(2)] the classification loss for the two detection heads, {\small$L_\mathrm{cls}^1(Y^*)$ and $L_\mathrm{cls}^2(Y^*)$,}
%defined as the cross-entropy loss {\small$\sum_i -y_i^*\log(p_i)$}.
%\item[(3)] the localization loss for the two detection heads, {\small$L_\mathrm{loc}^1(B^{**},Y^*)$} and {\small$L_\mathrm{loc}^2(B^{**},Y^*)$,}
%defined as the smooth L1 loss~\cite{fast-rcnn}.
%\end{itemize}

\section{Experiments}
\label{sec:experiment}

In this section,
we first introduce the benchmark datasets and implementation details.
Then,
we evaluate the stand-alone effect of CA-BBC.
Next,
we compare the proposed learning framework with state-of-the-art methods on multiple benchmarks.
Finally,
we conduct ablation study to dissect the proposed framework and provide qualitative results.

\subsection{Datasets}
Since most available datasets for object detection have been extensively verified by human annotators (an expensive process) and contain little noise,
we created noisy annotations using two popular benchmark datasets, PASCAL~\cite{pascal} and MS-COCO~\cite{mscoco},

First,
we generated synthetic noise to simulate human mistakes of different severity,
by corrupting the training annotation with a mixture of label noise and bounding box noise.
For label noise,
we follow previous works~\cite{mentornet,Arazo_ICML_2019} and generate symmetric label noise.
Specifically, we randomly choose $N_l\%$ of the training samples and change each of their labels to another random label. 
For bounding box noise,
we perturb the coordinates of all bounding boxes by a number of pixels uniformly drawn from $[-w N_b\%,+w N_b\%]$ ($w$ is bbox width)  for horizontal coordinates or $[-h N_b\%,+h N_b\%]$ ($h$ is bbox height)  for vertical coordinates.
We experiment with multiple combinations of label noise ranging from $0\%$ to $60\%$ and bounding box noise ranging from $0\%$ to $40\%$.
Under 40\% bbox noise, the average IoU between a noisy bbox and its corresponding clean bbox is only \textbf{0.45}.

%To enable our experiments,
%we corrupt the training data of two popular benchmark datasets, PASCAL VOC 2007 \& 2012~\cite{pascal} and MS-COCO~\cite{mscoco},
%with both label noise and bounding box noise.
%We define label noise level as $N_l$ and bounding box noise level as $N_b$.
%For label noise,
%we follow previous works~\cite{mentornet,Arazo_ICML_2019} and generate symmetric noise by randomly replacing the
%labels for $N_l\%$ of the training samples with other labels.
%For bounding box noise,
%we shift the coordinates of each bounding box by a number of pixels randomly drawn from $[-w N_b\%,+w N_b\%]$ ($w$ is bbox width)  for horizontal coordinates or $[-h N_b\%,+h N_b\%]$ ($w$ is bbox height)  for vertical coordinates.
%We experiment with multiple combinations of label noise ranging from $0\%$ to $60\%$ and bounding box noise ranging from $0\%$ to $40\%$.

PASCAL VOC contains 20 object classes.
We follow previous paper~\cite{faster_rcnn} and use the union set of VOC 2007 \textit{trainval} and VOC 2012 \textit{trainval} as our training data,
and VOC 2007 \textit{test} as our test data.
We report mean average precision (mAP@$.5$) as the evaluation metric.
MS-COCO contains 80 object classes,
we use \textit{train2017} as training data,
and report mAP@$.5$ and mAP@$[.5, .95]$ on \textit{val2017}.

We also mined large amounts of free training data with noisy annotations by using a poorly-trained detector to annotate unlabeled images.
%In order to evaluate the practical impact of the proposed method,
%we also create a dataset with machine-generated annotations.
We first train a Faster R-CNN detector on only $10\%$ of labeled data from COCO\textit{train2017},
which has a validation mAP@$.5$ of 40.5.
Then we use the trained detector to annotate \textit{unlabeled2017},
which contains 123k unlabeled images.
We use COCO\textit{unlabeled2017} with machine-generated annotations as our noisy training data.

\subsection{Implementation Details}
The proposed method is applicable to any two-stage detectors that are compatible with Precise RoI-Pooling~\cite{prroi}.
In our experiments,
we use the common Faster-RCNN~\cite{faster_rcnn} architecture with ResNet-50~\cite{resnet} and FPN~\cite{FPN} as the feature extractor.
We train the model using SGD with a learning rate of 0.02, a momentum of 0.9, and a weight decay of $1e-4$.
The hyper-parameters are set as $\lambda=0.1$, $T=0.4$, $\rho=0.5$, and $\alpha \in \{0,100,200\}$,
which are determined by the validation performance on 10\% of training data with clean annotations (only used for validation).
We implement our framework based on the mmdetection toolbox~\cite{mmdetection}.
In terms of computation time,
our method increases training time by ${\sim}24\%$ compared to vanilla training.
During inference,
we only use the first detection head unless otherwise specified,
which does not increase inference time.

\begin{table}[!t]
	\centering	
	\setlength\tabcolsep{5pt}
	%\resizebox{\columnwidth}{!}{%
	\begin{tabular}	{l | c | c | c | c | c | c }
		\toprule	 	
			BBox Noise & \multicolumn{3}{c|}{20\%}& \multicolumn{3}{c}{40\%}\\
			\midrule
			Label Noise  & 0\%& 20\%&40\%&0\%&20\%&40\%\\
			\midrule			
			\midrule
			vanilla~\cite{faster_rcnn,mmdetection} & 75.5 & 70.7  &66.9 & 59.3  &54.2 &50.0\\
			\midrule
			objectness maximization & 75.8 & 71.1  &67.4 & 63.2  &59.6 &55.7\\	
			\midrule
			 CA-BBC &\textbf{76.8} & \textbf{72.4} &\textbf{68.0}   & \textbf{67.8}  & \textbf{64.7}  &\textbf{61.7}\\

		\bottomrule
	\end{tabular}
	%}
	\caption
		{
		\small	
		Evaluation of class-agnostic bounding box correction.
		Numbers are mAP@$.5$ on PASCAL VOC 2007 test set for models trained with different mixtures of label noise and bbox noise.
		CA-BBC effectively disentangles bbox noise from label noise and correct it.
		}
	\label{tbl:bbox_agnostic}	
		\vspace{-4ex}
\end{table}		

%\begin{table*}[!t]
%	\centering	
%	\begin{tabular}	{l |c | c | c | c | c | c | c | c | c }
%		\toprule	 	
%			BBox Noise Level&\multicolumn{3}{c|}{0\%}&  \multicolumn{3}{c|}{20\%}& \multicolumn{3}{c}{40\%}\\
%			\midrule
%			Label Noise Level & 0\%& 20\%&40\%& 0\%& 20\%&40\%&0\%&20\%&40\%\\
%			\midrule			
%			Faster R-CNN & 78.2  & 72.9& 69.3&  75.5 & 70.7  &66.9 & 59.3  &54.2 &50.0\\
%			Faster R-CNN with CA-BBC &79.2& 73.6 & 69.9 &76.8 & 72.4 &68.0   & 67.8  & 64.7  &61.7\\	
%			
%			%Faster R-CNN with BBox Correction (alpha100)  & & 76.8 &67.2 & & 72.4 &  &  & 68.0 &59.2\\	
%			%Faster R-CNN with BBox Correction (alpha200)  & & 76.2 &67.8 & & & 64.7 &  & 67.7 &61.7\\		
%
%		\bottomrule
%	\end{tabular}
%	%}
%	\caption
%		{
%		\small	
%		mAP
%		}
%	\label{tbl:bbox_agnostic}	
%\end{table*}		

%\begin{table*}[!t]
%	\centering	
%	\begin{tabular}	{l |c | c | c | c | c | c | c | c | c }
%		\toprule	 	
%		Label Noise Level&\multicolumn{3}{c|}{0\%}&  \multicolumn{3}{c|}{20\%}& \multicolumn{3}{c}{40\%}\\
%		\midrule
%		BBox Noise Level & 0\%& 20\%&40\%& 0\%& 20\%&40\%&0\%&20\%&40\%\\
%		\midrule			
%		Faster R-CNN & 78.2  & 75.5& 59.3 &  72.9 & 70.7  &54.2 &69.3  &66.9 &50.0\\
%		Faster R-CNN with CA-BBC &79.2& 76.8 &67.8 &73.6 & 72.4 &64.7  & 69.9 & 68.0 &61.7\\	
%		
%		%Faster R-CNN with BBox Correction (alpha100)  & & 76.8 &67.2 & & 72.4 &  &  & 68.0 &59.2\\	
%		%Faster R-CNN with BBox Correction (alpha200)  & & 76.2 &67.8 & & & 64.7 &  & 67.7 &61.7\\		
%		
%		\bottomrule
%	\end{tabular}
%	%}
%	\caption
%	{
%		\small	
%		mAP
%	}
%	\label{tbl:bbox_agnostic}	
%\end{table*}		

\subsection{Evaluation on CA-BBC}

First,
we evaluate the effect of the proposed CA-BBC method by itself.
We train a detector following the proposed learning framework,
except that we only perform the first step of noise correction (\ie~CA-BBC).
Table~\ref{tbl:bbox_agnostic} shows the results on PASCAL VOC with different mixtures of label noise and bounding box noise.
Compared to vanilla training without any noise correction~\cite{faster_rcnn,mmdetection},
performing CA-BBC can significantly improve performance,
especially for higher level of bbox noise.
The improvement in mAP is consistent despite the increase of label noise,
which demonstrates the ability of CA-BBC to disentangle the two types of noise and effectively correct bbox noise.
We also demonstrate the effect of the proposed discrepancy minimization by removing $\mathcal{D}(p_1,p_2)$ from the loss in Eq.~\ref{eqn:Lb},
and only maximize the objectness of the bbox region,
which leads to lower performance.

\begin{figure*}[!t]
 \centering
   \includegraphics[width=\linewidth]{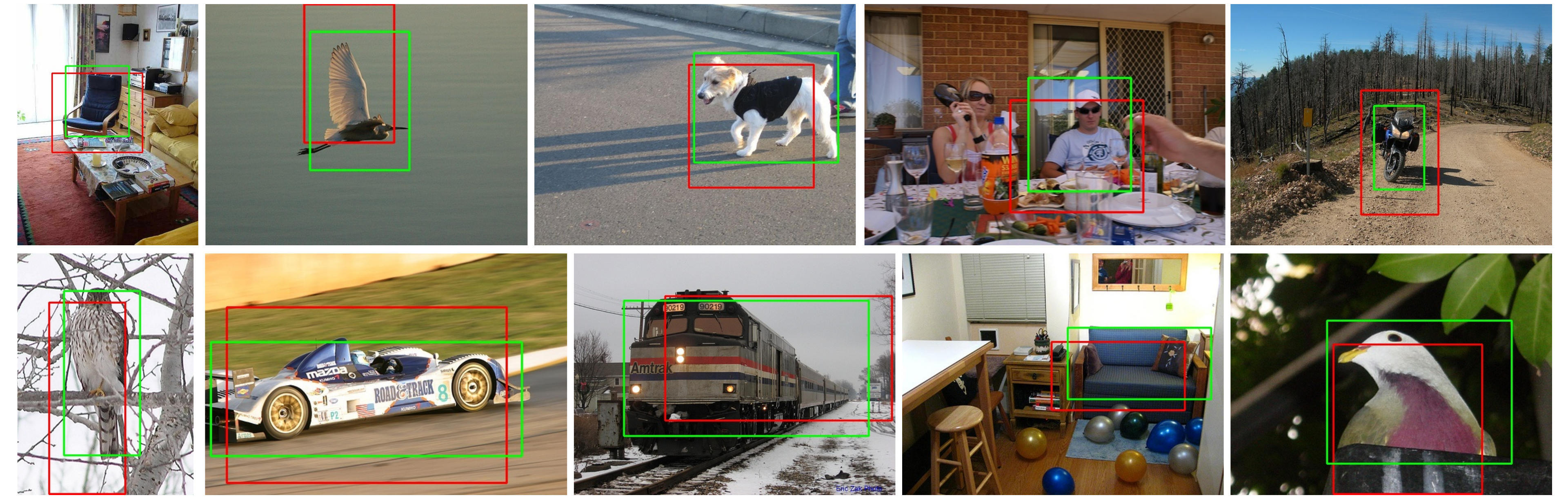}
   \vspace{-4ex}
   \caption
  	{
  	\small
		Examples of class-agnostic bounding box correction (CA-BBC) on PASCAL VOC dataset with 40\% label noise and 40\% bounding box noise.
		Noisy GT bounding boxes are in red and the corrected bounding boxes are in green.} 
  \label{fig:CA_example}
 \end{figure*}

\begin{figure}[!t]
 \centering
   \includegraphics[width=\linewidth]{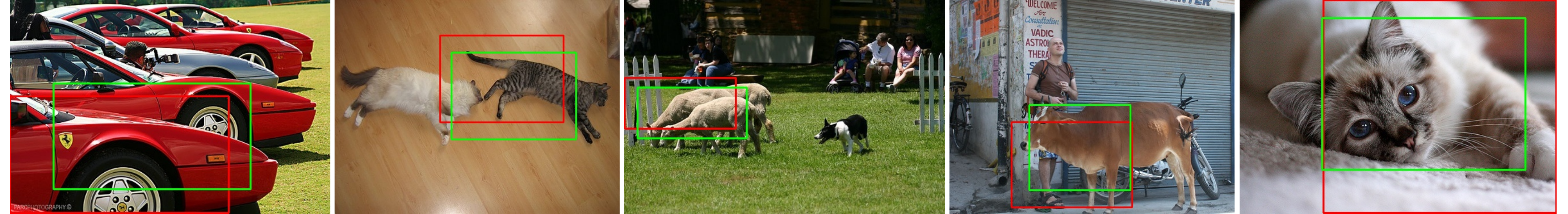}
   \caption
  	{
  	\small
		Examples of imperfect bbox corrections from CA-BBC. Noisy GT bounding boxes are in red and the corrected bounding boxes are in green.
		CA-BBC could be confused when the GT bboxes cover multiple object instances of the same class. It also sometimes fails to extend the box boundaries to cover the entire object.} 
  \label{fig:CA_fail_example}
  \vspace{-1ex}
 \end{figure}

We show some qualitative examples of CA-BBC in Figure~\ref{fig:CA_example}.
The noisy GT bboxes are shown in red whereas the corrected bboxes are shown in green.
CA-BBC can update the bounding boxes to more accurately capture the objects of interest.
We also provide some examples in Figure~\ref{fig:CA_fail_example} where CA-BBC fails to correct the bbox noise.
%Due to the objective function of CA-BBC (\ie~minimize classifier discrepancy and maximize region objectness),
It could be confused when the GT bboxes cover multiple object instances of the same class.
%or fails to extend the box boundaries to cover the entire object.

%\begin{table*}[!t]
%	\centering	
%	\begin{tabular}	{l |c | c | c | c | c | c | c | c | c |  c | c | c  }
%		\toprule	 	
%		Label Noise Level&\multicolumn{3}{c|}{0\%}&  \multicolumn{3}{c|}{20\%}& \multicolumn{3}{c|}{40\%}& \multicolumn{3}{c}{60\%}\\
%		\midrule
%		BBox Noise Level & 0\%& 20\%&40\%& 0\%& 20\%&40\%&0\%&20\%&40\%&0\%&20\%&40\% \\
%		\midrule	
%	    \midrule
%		Faster R-CNN & 78.2  & 75.5& 59.3 &  72.9 & 70.7  &54.2 &69.3  &66.9 &50.0 & 62.1 & 61.2 &45.9\\
%		
%		\midrule
%		Dual-Teacher + CA-BBC & & &    & & &    & &&69.0     & & &64.5\\		
%		
%		\bottomrule
%	\end{tabular}
%	%}
%	\caption
%	{
%		\small	
%		mAP
%	}
%	\label{tbl:pascal_result}	
%\end{table*}		

\begin{table*}[!t]
	\centering	
	\resizebox{\columnwidth}{!}{%
	\setlength\tabcolsep{3pt}
	\begin{tabular}	{l |c | c | c | c | c | c | c | c | c |  c | c | c  }
		\toprule	 	
		BBox Noise &\multicolumn{4}{c|}{0\%}&  \multicolumn{4}{c|}{20\%}& \multicolumn{4}{c}{40\%}\\
		\midrule
		Label Noise & 0\%& 20\%&40\%& 60\%& 0\%& 20\%&40\%& 60\%&0\%&20\%&40\%& 60\%\\
		\midrule	
	    \midrule
		 Vanilla~\cite{faster_rcnn,mmdetection} & 78.2  & 72.9& 69.3 & 62.1&  75.5 & 70.7  &66.9 & 61.2 & 59.3  &54.2 &50.0 & 45.9\\
		\midrule
		Co-teaching~\cite{co-teaching} &78.3 & 76.5&74.1 &69.9& 75.6& 73.2& 69.7&65.1 & 60.6& 59.7&55.8 &50.4 \\
		\midrule
		SD-LocNet~\cite{noisy_AAAI} &78.0 & 75.3& 73.0& 66.2& 75.3&72.1 & 67.5&64.0 & 59.7&58.7 &54.5 & 49.2\\
		\midrule
		NOTE-RCNN~\cite{noisy_ICCV} &78.6 &76.7 & 74.9& 69.9 & 76.0&73.7 & 70.1&65.8 & 63.4& 61.5&57.8 & 53.7\\
		\midrule
		
		Ours &\textbf{80.1} & \textbf{79.1} & \textbf{77.7}  & \textbf{74.1}& \textbf{77.9} & \textbf{76.7} &  \textbf{74.8} & \textbf{71.9}&\textbf{71.9} &  \textbf{70.6}  & \textbf{69.1} & \textbf{64.5}\\				
		
		\bottomrule
	\end{tabular}
	}
	\caption
	{
		\small	
		Comparison with state-of-the-art methods on PASCAL VOC dataset.
		Numbers indicate mAP@$.5$ on 2007 test set for models trained with different mixtures of label noise and bbox noise.
	}
	\label{tbl:pascal_result}	
	\vspace{-2ex}
\end{table*}

\subsection{Comparison with the State-of-the-arts}

We evaluate our full learning framework with two-step noise correction and compare it with multiple state-of-the-art methods for learning with noisy annotations.
%The baselines are described as follows.
We implement all methods using the same network architecture (\ie~Faster-RCNN with Precise RoI-Pooling).
Since previous methods operate in different settings as ours,
we adapt them for our problem to construct strong baselines as described in the following:

\begin{itemize}[leftmargin=*]\setlength\itemsep{0.5pt}
	\item Co-teaching~\cite{co-teaching} simultaneously trains two models where each model acts as a teacher for the other by selecting its small-loss samples within a mini-batch as clean data to train the other.
	We employ co-teaching into our dual-head network,
	where each detection head selects box samples with small classification loss to train the other head.
	Note that the RPN is trained on all boxes.   
	\item SD-LocNet~\cite{noisy_AAAI} proposes an adaptive sampling method that assigns a reliable weight to each box sample.
	Higher weights are assigned to samples with higher classification scores and lower prediction variance over consecutive training epochs.	
	\item NOTE-RCNN~\cite{noisy_ICCV} uses clean seed box annotations to train the bbox regression head.
	It also pretrains a teacher detector on the clean annotations for knowledge distillation.
	Because we do not have clean annotations,
	we follow previous works~\cite{co-teaching,Arazo_ICML_2019} and consider box samples with smaller classification loss as clean ones.
	We first train a detector in a standard manner to mine clean samples.
	Then we utilize the clean samples following NOTE-RCNN~\cite{noisy_ICCV}.
\end{itemize}

Table~\ref{tbl:pascal_result} shows the comparison results on PASCAL VOC dataset, where the training data contains different mixtures of label noise and bbox noise. Our method significantly outperforms all other methods across all noise settings. For high levels of noise ($N_b=40\%,N_l\in\{40\%,60\%\}$), our method achieves ${\sim}20\%$ improvement in mAP compared to vanilla training, and ${>}10\%$ improvement compared to the state-of-the-art NOTE-RCNN~\cite{noisy_ICCV}, which demonstrates the advantage of the proposed two-step noise correction method for dealing with entangled annotation noise. 

On clean training data with $0\%$ annotation noise,
our method can still improve upon vanilla training by ${+}1.9\%$ in mAP. 
The improvement mostly comes from using the proposed soft labels,
for two reasons.
First, compared to the one-hot GT labels,
soft labels contain more information about an image region in cases where multiple objects co-exists in the same bounding box,
which often occur in the training data (\eg~the bounding box of a bike would also also cover a person if the person is standing in front of the bike).
In such scenarios, using hard labels would undesirably penalize the scores for all non-GT classes, even if such classes exist in the box.
Moreover,
using soft labels has the effect of label smoothing,
which could prevent overfitting and improve a model's generalization performance~\cite{smooth}.
%In the clean training data,
%bounding boxes could intersect with each other (\eg a person standing in front of a car),
%hence a box could contain parts of other objects besides its corresponding GT object. 
%In such cases,
%using the one-hot GT labels would undesirably penalize the scores for all non-GT classes.
%However,
%using the proposed soft labels would consider the presence of other objects in the box and reduce the penalty on those classes,
%which leads to better training of the classifier. 

\begin{table*}[!t]
	\setlength\tabcolsep{1pt}
	\centering	
		\resizebox{\columnwidth}{!}{%
	\begin{tabular}	{l |c  c| c  c |c  c }
		\toprule	 	
		\multirow{2}{*}{Method}  & \multicolumn{2}{c|}{$N_b=20\%,N_l=20\%$}& \multicolumn{2}{c|}{$N_b=40\%,N_l=40\%$}&
		\multicolumn{2}{c}{machine-generated}\\
		\cmidrule{2-7}
		&mAP@$.5$~ &mAP@$[.5, .95]$ & mAP@$.5$~ &mAP@$[.5, .95]$&mAP@$.5$~ &mAP@$[.5, .95]$\\
		\midrule	
	    \midrule
	    Vanilla~\cite{faster_rcnn,mmdetection} &47.9& 23.9 & 29.7 & 10.3& 41.5 &21.5\\
	    \midrule
		Co-teaching~\cite{co-teaching}& 49.7 & 24.6  & 35.9&14.6& 41.4&21.5\\
		\midrule
		SD-LocNet~\cite{noisy_AAAI} & 49.3 &24.5  &35.1 &13.9&42.8&21.9\\
		\midrule
		NOTE-RCNN~\cite{noisy_ICCV} & 50.4 &25.1  &38.5 &15.2& 43.1&22.0\\
		\midrule		
		Ours &\textbf{53.5} & \textbf{27.7}  &\textbf{47.4} &\textbf{21.2}&\textbf{46.5} &\textbf{23.2}\\	
		\midrule
		Clean & 54.5&31.4		& 54.5&31.4	& 54.5&31.4	\\
		
		\bottomrule
	\end{tabular}
	}
	\caption
	{
		\small	
		Comparison with state-of-the-art methods on MS-COCO dataset.
		We report the mAP on val2017 for methods trained with different annotation noise.
	}
	\label{tbl:coco_result}	
	\vspace{-2ex}
\end{table*}		
%
%
%\begin{table*}[!t]
%	\centering	
%	\setlength\tabcolsep{5pt}
%	\resizebox{\textwidth}{!}{%
%	\begin{tabular}	{l |  l |c |c |c | c |c | c }
%		\toprule	 	
%		Noise &Method   & Vanilla~\cite{faster_rcnn,mmdetection} & 	Co-teaching~\cite{co-teaching} &SD-LocNet~\cite{noisy_AAAI} & NOTE-RCNN~\cite{noisy_ICCV} & Ours & Clean\\
%			
%	    \midrule
%	    \multirowcell{2}[0pt][l]{$N_l=20\%$\\$N_b=20\%$} & 
%	    mAP@$.5$  & 47.9 &49.7 &49.3&50.4& \textbf{53.5} & 54.5\\
%	    &mAP@$[.5, .95]$ &  23.9 & 24.6 &24.5&25.1& \textbf{27.7} &  31.4\\
%	    \midrule
%	    \multirowcell{2}[0pt][l]{$N_l=40\%$\\$N_b=40\%$} & 
%		mAP@$.5$  & 29.7 &35.9 &35.1&38.5& \textbf{47.4} & 54.5\\
%		&mAP@$[.5, .95]$ &  10.3 &14.6 &13.9&15.2& \textbf{21.2} &  31.4\\	
%		
%		\bottomrule
%	\end{tabular}
%	}
%	\caption
%	{
%		\small	
%		Comparison with state-of-the-art methods on MS-COCO dataset.
%		We report the mAP on val2017.
%		%The training data contains $20\%$ label noise and $20\%$ bbox noise. 
%	}
%	\label{tbl:coco_result}	
%\end{table*}		

\def\arraystretch{1.2}
\newcolumntype{?}{!{\vrule \hspace{0.2pt} \vrule}}
\begin{table*}[!t]
	\resizebox{\columnwidth}{!}{%
\setlength\tabcolsep{3pt}
	\centering	
	%\resizebox{\textwidth}{!}{%
	\begin{tabular}	{ |c | c | c | c  ? c | c | c | c | c | c | c | c | c| c  }
		\cline{1-13}	 	
		\multirowcell{2}[0pt][l]{Forward\\Corr.} &
	\multirowcell{2}[0pt][l]{Dual\\Heads}  & \multirowcell{2}[0pt][l]{CA\\BBC} & \multirowcell{2}[0pt][l]{Dual\\Infer.} &  \multicolumn{3}{c|}{0\%} & \multicolumn{3}{c|}{20\%}& \multicolumn{3}{c|}{40\%} &{\footnotesize \hspace{-1ex} ($N_b$)}\\
		\cline{5-13}
	
	 & &   &   &20\%&40\%& 60\%&20\%&40\%& 60\%&20\%&40\%& 60\% &{\footnotesize \hspace{-1ex} ($N_l$)}\\

		\cline{1-13}
		\checkmark	 &  &    & &     78.9 & 77.4&  73.4&   75.9& 73.6& 68.8&   67.2 &  65.3 &59.1 &\\		
		\cline{1-13}
		\checkmark	 &\checkmark &      &   &   79.1& 77.7& 74.1 & 76.2   & 74.1 & 69.8 &  67.9  & 66.0  &60.3 &\\		
		\cline{1-13}
		\checkmark	 &\checkmark & \checkmark &    &  79.1& 77.7& 74.1 &  76.7  &  74.8 & 71.9&  70.6  &  69.1 & 64.5 &\\				
		\cline{1-13}
	    \checkmark	 &\checkmark & \checkmark &  \checkmark     &     \textbf{79.6}& \textbf{78.3}& \textbf{74.8}&  \textbf{77.3}  & \textbf{75.2} & \textbf{72.5}&  \textbf{71.3}  &  \textbf{69.8} & \textbf{65.4} &\\				
	    %\hline
	    %\checkmark & \checkmark & \checkmark & \checkmark       & 79.4&  77.9& 74.6  & 76.9   &75.4 &72.4 &  70.7  &  69.7 &64.9\\		
	    %\hline
	   % \checkmark & \checkmark & \checkmark & \checkmark   &  \checkmark &  &    & & &    &   &\\		 
		\cline{1-13}
	\end{tabular}%
	}
	\caption
	{
		\small	
		Ablation study to examine the effect of each component in the proposed framework.
		Numbers indicate mAP@$.5$ on PASCAL VOC 2007 test set.
		The results validate the efficacy of the proposed CA-BBC and dual-head noise correction method.
		Ensemble of the two detection heads during inference can further boost performance. 		
	}
	\label{tbl:ablation}	
	\vspace{-2ex}
\end{table*}

Table~\ref{tbl:coco_result} shows the results on MS-COCO dataset.
Our method outperforms all state-of-the-art baselines by a large margin.
Under $40\%$ of label and bbox noise,
vanilla training results in a catastrophic degradation of ${-}24.8\%$ in mAP@$.5$ compared to training on clean data,
whereas the proposed method can reduce the performance drop (from clean) to an acceptable ${-}7.1\%$.
The proposed method also achieves improvement under machine-generated noise,
which validates its practical usage to train detectors by utilizing free unlabeled data.

\subsection{Ablation Study}

In this section, we conduct ablation studies to dissect and analyse the proposed framework. We also provide qualitative results to demonstrate the effectiveness of the proposed dual-head noise correction. 

\begin{figure}[!t]
	\centering
	\begin{minipage}{0.47\textwidth}
	\centering
	\includegraphics[width=\linewidth]{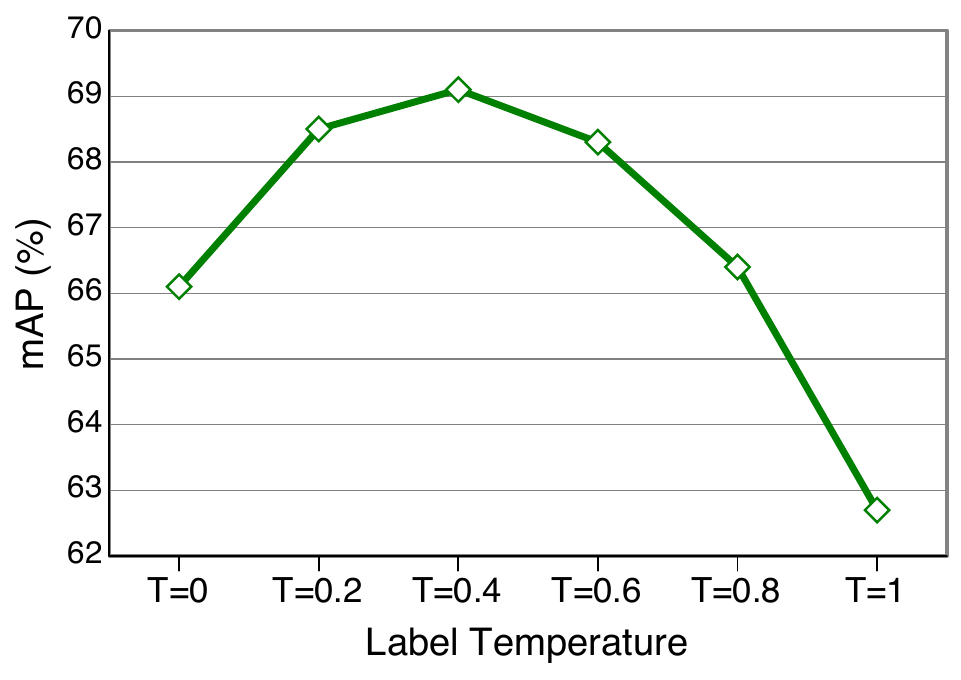}
	(a)
\end{minipage}
\begin{minipage}{0.5\textwidth}
	\centering
	\includegraphics[width=\linewidth]{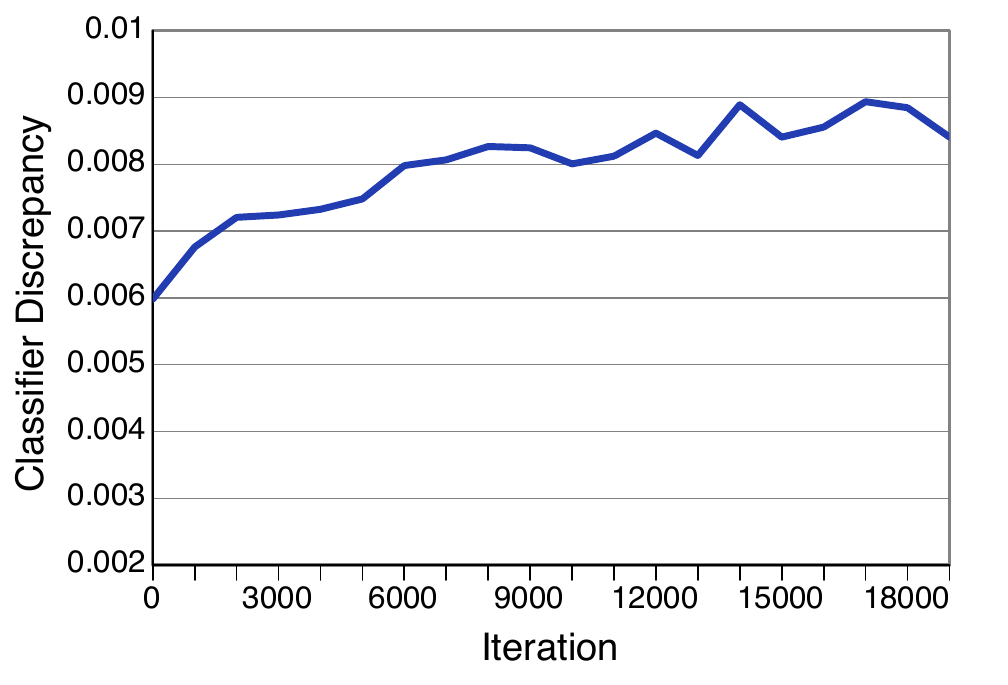}
	(b)
\end{minipage}
	\vspace{-2ex}
   \caption
  	{
  	\small
		(a) Test performance with different sharpening temperature $T$ for soft labels (Eq.~\ref{eqn:sharp}).
		$T$ controls the entropy of the label distribution.
		%Training data is PASCAL VOC with 40\% label noise and 40\% bbox noise.
		(b) Divergence between the two detection heads measured by their classifier discrepancy (Eq.~\ref{eqn:dis}), averaged over 1000 iterations (mini-batches).		
		The two heads can stay sufficiently diverged due to different initialization and different proposal sampling during training.
Training data for both figures is PASCAL VOC with 40\% label noise and 40\% bbox noise.		
	} 
  \label{fig:temperature}
 \end{figure}

\vspace{1ex}
\noindent\textbf{Model variants.}
We add or drop different components in our framework to examine their effects. Table~\ref{tbl:ablation} shows the ablation study results on PASCAL VOC datatset. 
%Below we explain the results in detail.

\vspace{-1ex}
\begin{itemize}[leftmargin=*]\setlength\itemsep{0pt}
\item In the first row, we perform forward noise correction with only one detection head,
by using its output to create soft labels and regress bounding boxes.
Compared with the proposed dual-head network where knowledge is distilled from the ensemble,
using a single head suffers from the confirmation bias problem where the model's prediction error would accumulate~\cite{Tarvainen_NIPS_17,co-teaching} and thus degrade the performance.

\item In the second row,
we remove CA-BBC from the proposed framework and only perform the dual-head noise correction (Section~\ref{sec:dual-head}).
Compared with the results using the proposed two-step noise correction (the third row),
the performance decreases considerably for higher level (40\%) of bounding box noise, 
which validates the importance of the proposed CA-BBC.
\item The third row shows the results using the proposed method.
\item In the last row,
we use the ensemble of both detection heads during inference by averaging their outputs,
which leads to further performance improvement.	 
\end{itemize}

\begin{figure*}[!t]
 \centering
   \includegraphics[width=\linewidth]{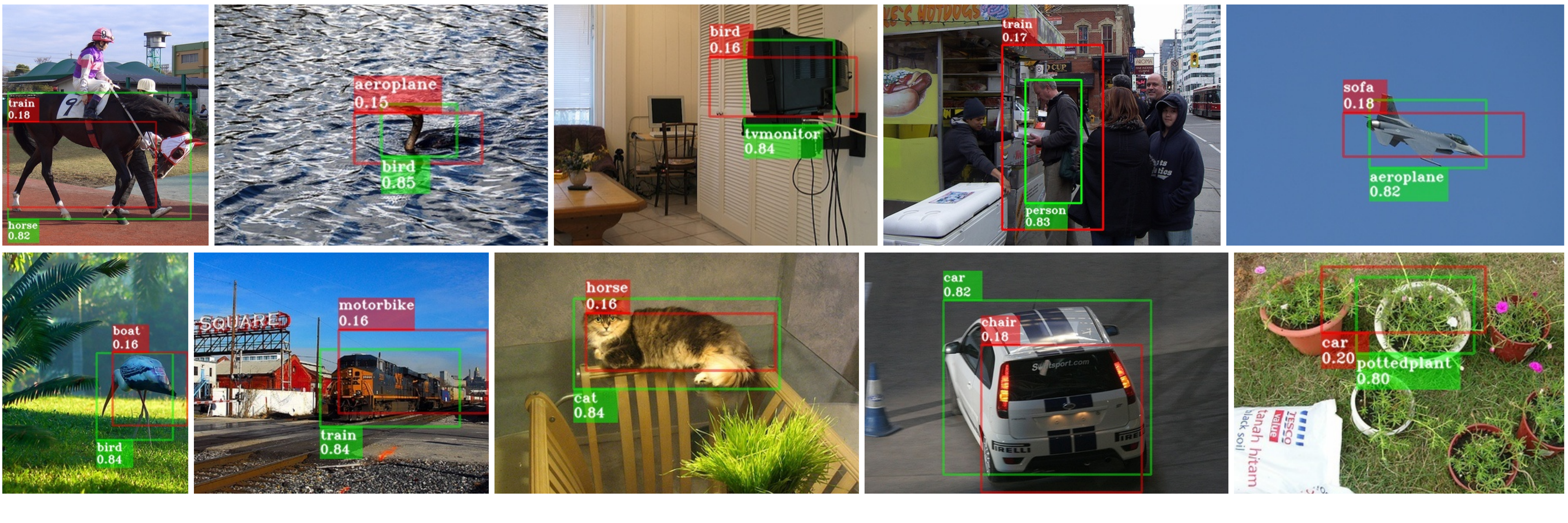}
   \vspace{-4ex}
   \caption
  	{
  	\small
		Examples of dual-head noise correction on PASCAL VOC dataset with 40\% label noise and 40\% bounding box noise (best viewed in color with zoom in).
		Noisy GT labels and GT bounding boxes are shown in red,
		whereas the corrected labels and bounding boxes are shown in green.
	    Numbers under the class names indicate corresponding values in the soft labels.} 
  \label{fig:dual_example}
 \end{figure*}
\noindent\textbf{Soft label temperature.}
In Figure~\ref{fig:temperature}(a),
we vary the sharpening temperature $T$,
which controls the entropy of the soft label distribution (Eq.~\ref{eqn:sharp}).
A lower temperature encourages the model to produce more confident predictions,
whereas a higher temperature emphasizes more on the uncertainty of the label.
$T\rightarrow0$ produces one-hot labels,
which corresponds to the hard bootstrapping method~\cite{Reed_2015_ICLR},
whereas $T=1$  corresponds to the soft bootstrapping method~\cite{Reed_2015_ICLR}.
We observe that acceptable values of $T$ ranges around $0.2\sim0.6$.
We use $T=0.4$ in all experiments.

\vspace{1ex}
\noindent\textbf{Dual-head divergence analysis.}
In Figure~\ref{fig:temperature}(b),
we show that the two detection heads can stay sufficiently diverged during the entire training process,
due to different parameter initialization and different region proposal sampling.
We measure divergence using classifier discrepancy (Eq.~\ref{eqn:dis}).
Keeping the two detection heads diverged is important for several reasons.
First, the effectiveness of the proposed CA-BBC requires disagreement between the two classifiers.
Second, diverged detection heads can filter different kinds of noise,
which results in more effective dual-head noise correction.
Furthermore,
diverged dual heads can lead to better ensemble performance during test.

\vspace{1ex}
\noindent\textbf{Noise correction examples.}
In Figure~\ref{fig:dual_example},
we show example images with annotations before and after the proposed two-step noise correction method.
Noisy GT labels and bounding boxes are shown in red whereas the corrected labels and boxes are shown in green.
The proposed method can effectively clean the annotation noise by
1) regressing the bounding boxes to more accurate locations and 2) assigning high soft label values to the correct classes.

	\section{Conclusion}
\label{sec:conclusion}
To conclude, this paper addresses a new challenging research problem, which aims to train accurate object detectors from noisy annotations that contain entangled label noise and bounding box noise. We propose a noise-resistant learning framework which jointly optimizes noisy annotations and model parameters. A two-step noise correction method is proposed, where the first step performs class-agnostic bbox correction to disentangle bbox noise and label noise, and the second step performs dual-head noise correction by self-distillation. Experiments on both synthetic noise and machine-generated noise validate the efficacy of the proposed framework over state-of-the-art methods and variants of itself. We believe that our work is one step forward towards alleviating human from the tedious annotation effort. In the future, we plan to extend the proposed framework to segmentation tasks, as well as conducting human study to examine the effect of noise-resistant training for human-in-the-loop learning.
	\clearpage
	\bibliographystyle{splncs04}
	\bibliography{bib}

\end{document}